\documentclass[letterpaper, 10 pt, conference]{ieeeconf}  %

\IEEEoverridecommandlockouts                              %

\usepackage{silence}
\usepackage{amsmath}
\usepackage{amsfonts}
\usepackage{amssymb}
\usepackage{graphicx}
\usepackage{color}
\usepackage{units}
\usepackage[style=base]{subcaption}

\usepackage{amsthm}

\usepackage[hidelinks]{hyperref}
\usepackage[capitalise]{cleveref}

\DeclareCaptionLabelSeparator{periodspace}{.\quad}
\theoremstyle{plain}

\crefname{equation}{}{} %
\crefname{section}{Sec.}{Sec.}

\newcommand{\E}{\mathrm{E}}             %
\newcommand{\T}{\mathrm{T}}             %
\newcommand{\diag}{\mathrm{diag}}       %
\newcommand{\argmax}{\operatornamewithlimits{argmax}}
\newcommand{\argmin}{\operatornamewithlimits{argmin}}

\newcommand{\ours}{MF-ES } %

\newcommand{\mb}[1]{\mathbf{#1}}        %
\newcommand{\bs}[1]{\boldsymbol{#1}}    %

\newcommand{\exper}{\mathrm{exp}}
\newcommand{\simu}{\mathrm{sim}}
\newcommand{\error}{\mathrm{err}}

\graphicspath{{figures/}}

\title{\LARGE \bf
Virtual vs. Real: Trading Off Simulations and Physical Experiments \\ in Reinforcement Learning with Bayesian Optimization
}

\author{Alonso Marco$^{1,5}$, Felix Berkenkamp$^{2,5}$, Philipp Hennig$^{1,5}$, Angela P. Schoellig$^{3}$,\\ Andreas Krause$^{2,5}$, Stefan Schaal$^{1,4,5}$, and Sebastian Trimpe$^{1,5}$ %
\thanks{$^{1}$ Max Planck Institute for Intelligent Systems, T\"{u}bingen, Germany. Email: \mbox{\textless firstname\textgreater.\textless lastname\textgreater @tuebingen.mpg.de}}%
\thanks{$^{2}$  Learning \& Adaptive Systems Group (LAS), Department of Computer Science, ETH Zurich, Switzerland. Email: \mbox{\{befelix, krausea\}@ethz.ch}}%
\thanks{$^{3}$  Dynamic Systems Lab (DSL), University of Toronto Institute for Aerospace Studies (UTIAS), Canada. Email: \mbox{schoellig@utias.utoronto.ca}}%
\thanks{$^{4}$  Computational Learning and Motor Control Lab, University of Southern California, USA.}%
\thanks{$^{5}$  Max Planck ETH Center for Learning Systems, T\"ubingen, Germany, and Z\"urich, Switzerland.}%
\thanks{This research was supported in part by the Max Planck ETH Center for Learning Systems, the Max Planck Society, SNSF grant {200020\_159557}, NSERC grant {RGPIN-2014-04634}, the Connaught New Researcher Award, and National Science Foundation grants IIS-1205249, IIS-1017134, EECS-0926052, the Office of Naval Research, the Okawa Foundation}%
}%

\usepackage{fancyhdr}
\newcommand{\mytitle}{\textbf{Accepted final version.}
To appear in \textit{2017 IEEE International Conference on Robotics and Automation}.\\
\copyright 2017 IEEE. Personal use of this material is permitted. Permission from IEEE must be obtained for all other uses, in any current or future media, including reprinting/republishing this material for advertising or promotional purposes, creating new collective works, for resale or redistribution to servers or lists, or reuse of any copyrighted component of this work in other works.}
\begin{document}

\maketitle
\thispagestyle{fancy}	%
\pagestyle{empty}

\begin{abstract}
In practice, the parameters of control policies are often tuned manually. This is time-consuming and frustrating. Reinforcement learning is a promising alternative that aims to automate this process, yet often requires too many experiments to be practical. In this paper, we propose a solution to this problem by exploiting prior knowledge from simulations, which are readily available for most robotic platforms. Specifically, we extend Entropy Search, a Bayesian optimization algorithm that maximizes information gain from each experiment, to the case of multiple information sources. The result is a principled way to automatically combine cheap, but inaccurate information from simulations with expensive and accurate physical experiments in a cost-effective manner. We apply the resulting method to a cart-pole system, which confirms that the algorithm can find good control policies with fewer experiments than standard Bayesian optimization on the physical system only.
\end{abstract}

\section{Introduction}
\label{sec:introduction}
Typically, the control policies that are used in robotics depend on a small set of tuning parameters. To achieve the best performance on the real system, these parameters are usually tuned manually in experiments on the physical platform. Policy search methods in reinforcement learning aim to automate this process~\cite{Sutton1998Reinforcement}. However, without prior knowledge, these methods can require significant amounts of experimental time before determining optimal, or even only reasonable parameters. In robotics, simulation models of the robotic system are usually available, e.g., as a by-product of the design process. While exploiting knowledge from simulation models has been considered before, no principled way to trade off between the relative costs and accuracies of simulations and experiments exists~\cite{kober2013reinforcement}. As a result, state-of-the-art reinforcement learning methods require more experimental time on the real system than necessary.

\begin{figure}[t]
  \vspace{1.5mm}
  \includegraphics[scale=1]{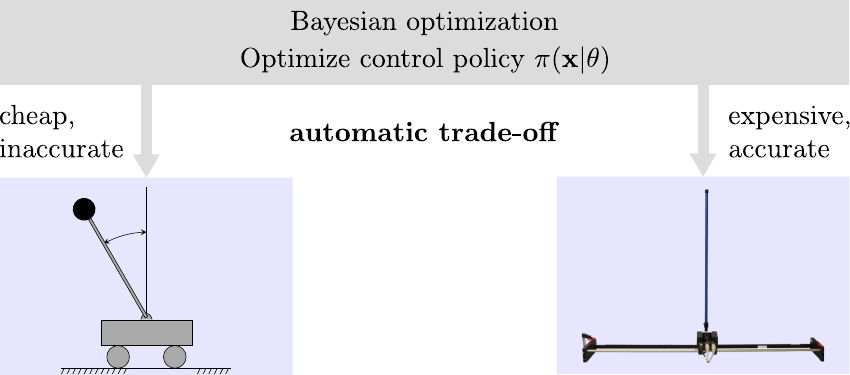}
  \caption{The proposed algorithm optimizes the parameters $\bs{\theta}$ of a control policy based on data of a cheap, but inaccurate simulation and expensive data from the real system. By actively trading off between the information that can be gained from each system relative to their costs, the algorithm requires significantly fewer evaluations on the physical system.}
  \label{fig:concept}
\end{figure}

In this paper, we propose a new reinforcement learning method that can automatically optimize the parameters of control policies based on data from different information sources, such as simulations and experiments. Specifically, we use an extension of Entropy Search~\cite{Hennig2012Entropy,Villemonteix2008Informational}, a Bayesian optimization framework for information-efficient global optimization. The resulting method automatically trades off the amount of information gained from different sources with their respective costs and requires fewer physical experiments to determine optimal parameters (see  \cref{fig:concept}).

\textbf{Related work:} Improving the performance of reinforcement learning with prior model information from a simulator has been considered before. A typical approach is two-stage learning, where algorithms are trained for a certain amount of time in simulation in order to warm-start the learning on the real robot~\cite{kober2013reinforcement}. For example,~\cite{Cutler2015Efficient} reports performance improvements when using model information from simulation as a prior for real experiments. Transfer learning is a similar approach that aims to generalize between different tasks, rather than from a simulated model to the real system~\cite{Taylor2009Transfer}. The work in~\cite{Cutler2015Realworld} learns an optimal policy and value function of a finite Markov decision process based on models with different accuracies. They rely on hierarchical models and switch to higher accuracy models once a threshold accuracy has been reached at a lower level.
A commonly used reinforcement learning method is policy gradients~\cite{Peters2006Policy}, where policy parameters are improved locally along the gradient. In this setting, simulation knowledge can be used to estimate the gradient of real experiments~\cite{Abbeel2006Using}. However, policy gradient methods only converge to locally optimal parameters.
None of the above methods explicitly considers the cost of experiments on a robot. In this paper, we actively trade off the different costs and information gains associated with simulation and real experiments and obtain globally optimal parameter estimates.

A method that has been particularly successful for para\-meter optimization in robotics is Bayesian optimization~\cite{Mockus1989Bayesian}. In particular, methods based on Gaussian process (GP,~\cite{Rasmussen2006Gaussian}) models are widely used because of their ability to determine globally optimal parameters within few evaluations. Examples include gait optimization in legged robots~\cite{Lizotte2007Automatic} and controller optimization for a snake-like robot~\cite{Tesch2011Using}. In~\cite{Marco2016Automatic}, the controller parameters of a linear state-feedback controller were optimized using the LQR framework as a low-dimensional representation of controller policies, while in~\cite{Abdelrahman2016Bayesian} the control policy itself was defined by Bayesian optimization with a specifically chosen kernel. Safety constraints on the robot during the optimization process were considered in~\cite{Berkenkamp2016Bayesian}. A comparison of different Bayesian and non-Bayesian global optimization methods can be found in~\cite{Calandra2014Experimental}. All the previous methods use Bayesian optimization directly on the real system. In contrast, we consider an extension of Bayesian optimization that can extract additional information from a simulator and speed up the optimization process.

The methodology herein is related to multi-task Bayesian optimization, where one aims to transfer knowledge about two related tasks~\cite{Krause2011Contextual,Swersky2013MultiTask}. A GP model with multiple information sources was first considered in~\cite{Kennedy2000Predicting}. Since then, optimization with multiple information sources has been considered under strict requirements, such as models forming a hierarchy of increasing accuracy and without considering different costs~\cite{Forrester2007Multifidelity,Kandasamy2016Multifidelity}. More recently,~\cite{Poloczek2016MultiInformation} used a myopic policy, called the `knowledge gradient' by~\cite{Frazier2009KnowledgeGradient}, in order determine, which parameters to evaluate.

\textbf{Our contribution:} In this paper, we present a Bayesian optimization algorithm for multiple information sources.
We use entropy~\cite{Hennig2012Entropy,Villemonteix2008Informational} to measure the information content of simulations and experiments. Since this is an appropriate unit of measure for the utility of both sources, our algorithm is able to compare physically meaningful quantities in the same units on either side, and trade off accuracy for cost. As a result, the algorithm can automatically decide whether to evaluate cheap, but inaccurate simulations or perform expensive and precise real experiments. We apply the method to optimize the policy of a cart-pole system and show that this approach can speed up the optimization process significantly compared to standard Bayesian optimization~\cite{Marco2016Automatic}. The main contributions of the paper are \textit{(i)} a novel Bayesian optimization algorithm that can trade off between costs of multiple information sources and \textit{(ii)} the first application of such a framework to the problem of reinforcement learning and optimization of controller parameters.

For convenience within the next sections, we rename the concepts \emph{accuracy} and \emph{cost}: We now refer to the lack of accuracy of a controller as \emph{cost}. Furthermore, we now denominate the cost of retrieving an evaluation from a specific information source as \emph{effort}.

\section{Problem Statement}
\label{sec:problem_statement}

We consider a reinforcement learning setting, where we aim to find an optimal policy to complete a certain task on a dynamic system. While we do not have access to a perfect model of the system, we assume that a control policy is available, which is parameterized by parameters~$\bs{\theta}$ within some domain~${\mathcal{D}}$. The goal is to determine the optimal parameters~$\bs{\theta}_{\mathrm{min}}$ that globally minimize the cost of a task,
\begin{equation}
  \bs{\theta}_\mathrm{min} \in \argmin_{\bs{\theta} \in \mathcal{D}} J(\bs{\theta}).
  \label{eq:cost_function}
\end{equation}
The cost~$J(\bs{\theta})$ measures the performance of the policy on a certain task on the real system. For example, one evaluation of the cost function could consist of controlling a robot with the parameterized policy and measuring an error signal over a fixed time horizon. To solve the optimization problem in~\cref{eq:cost_function}, we can query a parameter vector~$\bs{\theta}_n$ at each iteration~$n$ and observe the resulting performance~$J(\bs{\theta}_n)$. Since these experiments cause wear in the robot and take time, the goal is to minimize the number of iterations before the optimal parameters in~\cref{eq:cost_function} are determined.

We assume that a simulation of the system is available, which we want to exploit to solve~\eqref{eq:cost_function} more efficiently with fewer evaluations on the physical system. While simulations are only an approximation of the real world and cannot be used to determine the optimal parameters in~\cref{eq:cost_function} directly, they can be used to provide an estimate~$J_\mathrm{sim}(\bs{\theta})$ of the true cost. We use this estimate to obtain information about the location of the optimal parameters on the real system.
As a result, at each iteration~$n$, we do not only choose the next parameters~$\bs{\theta}_n$ to evaluate, but also whether to perform a simulation or an experiment.

Both experiments, in the real world and in simulation, have physically meaningful evaluation efforts associated to them. For example, the effort may account for the amount of time required to complete an experiment/simulation and for monetary costs such as wear in the system. The overall goal is to minimize the total effort incurred in the experiments and simulations until the optimal parameters~\cref{eq:cost_function} on the real system are found.

\section{Preliminaries}
\label{sec:preliminaries}

We start by introducing the necessary background information on GPs and Bayesian optimization.

\subsection{Gaussian Processes (GPs)}
\label{sec:background}

While the cost~$J(\bs{\theta})$ in~\cref{eq:cost_function} can easily be evaluated in an experiment for a given parameter~$\bs{\theta}$, the functional relationship between parameters and the cost is unknown \textit{a priori}. We use GPs as a nonparametric model to approximate the unknown function. The goal is to find an approximation of the nonlinear map,~${J(\bs{\theta})\colon \mathcal{D} \mapsto \mathbb{R}}$, from an input vector~${ \bs{\theta} \in \mathcal{D} }$ to the function value~$J(\bs{\theta})$. This is accomplished by modeling function values $J(\bs{\theta})$, associated with different values of $\bs{\theta}$, as random variables so that any finite number of these random variables have a joint Gaussian distribution~\cite{Rasmussen2006Gaussian}.

For the nonparametric regression, we define a prior mean function~$m(\bs{\theta})$, which encodes prior knowledge about the function~$J(\cdot)$, and a covariance function~$k(\bs{\theta}, \bs{\theta}')$, which defines the covariance of any two function values,~$J(\bs{\theta})$ and~${J(\bs{\theta}')}$, ${\bs{\theta}, \bs{\theta}' \in \mathcal{D}}$, and is used to model uncertainty about the mean estimate. The latter is also known as the kernel. The choice of kernel is problem-dependent and encodes assumptions about smoothness and rate of change of the unknown function,~$J(\cdot)$.

The GP framework can be used to predict the function value~${J(\bs{\theta}^*)}$ at an arbitrary input~${ \bs{\theta}^* \in \mathcal{D} }$,  based on a set of~$n$ past observations ${\mathcal{D}_n = \{ \bs{\theta}_i, \hat{J}(\bs{\theta}_i) \}_{i=1}^n}$. We assume that observations are noisy measurements of the true function; that is,
\begin{equation}
	\hat{J}(\bs{\theta}) = J(\bs{\theta}) + \omega(\bs{\theta}),
	\label{eq:j_measurement}
\end{equation}
where the noise ${\omega(\bs{\theta}) \sim \mathcal{N}(0,\eta^2(\bs{\theta}))}$ depends on the input. Conditioned on the previous observations, the mean and variance of the posterior normal distribution are
\begin{align}
	\mu_n(\bs{\theta}^*) &= m(\bs{\theta}^*) + \mb{k}_n(\bs{\theta}^*)  \mb{K}_n^{-1} \hat{\mb{y}}_n,
	\label{math:gp_prediction_mean} \\
	\sigma^2_n(\bs{\theta}^*) &= k(\bs{\theta}^*,\bs{\theta}^*) - \mb{k}_n(\bs{\theta}^*) \mb{K}_n^{-1} \mb{k}_n^\T(\bs{\theta}^*),
	\label{math:gp_prediction_variance}
\end{align}
where~${
	\hat{\mb{y}}_n = \left[
	\begin{matrix}
		\hat{J}(\bs{\theta}_1) - m(\bs{\theta}_1),\dots,\hat{J}(\bs{\theta}_n) - m(\bs{\theta}_n)
	\end{matrix}
	\right] ^\T
}$
is the vector of observed, noisy deviations from the mean,
the vector
${\mb{k}_n(\mb{a}^*) =
\left[ \begin{matrix}
	k(\bs{\theta}^*,\bs{\theta}_1),\dots,k(\bs{\theta}^*,\bs{\theta}_n)
\end{matrix}  \right]}$
contains the covariances between the new input~$\bs{\theta}^*$ and the observed data points in~$\mathcal{D}_n$, and
the symmetric matrix~${\mb{K}_n \in \mathbb{R}^{n \times n}}$ has entries ${[\mb{K}_n]_{(i,j)} = k(\bs{\theta}_i, \bs{\theta}_j) + \delta_{ij} \eta^2(\bs{\theta}_i) }$, ${i,j\in\{1,\dots,n\}}$.

\subsection{Bayesian Optimization}
\label{sec:bayesian_optimization}

We want to use the GP model of the cost function for parameter optimization. Using statistical models of an objective function for optimization is known as Bayesian optimization~\cite{Mockus1989Bayesian} in the literature. It comprises a class of data-efficient optimization methods that aim to determine the global optimum of cost functions that are expensive to evaluate. In our case, each evaluation of the cost with certain controller parameters on the robot cause wear in the system and may take a long time to perform. In the case of GP models, the mean,~\cref{math:gp_prediction_mean}, and variance,~\cref{math:gp_prediction_variance} can be used to determine new parameters to evaluate that are promising candidates for the global optimum. For example,~\cite{Srinivas2010Gaussian} uses upper confidence bounds that allow for provable convergence guarantees.

In this paper, we build upon the Entropy Search (ES,~\cite{Hennig2012Entropy}) algorithm, which selects parameters in order to maximally reduce the uncertainty about the location of the minimum of~$J(\bs{\theta})$ in each step. It quantifies this uncertainty through the entropy of the distribution over the location of the minimum,
\begin{equation}
	p_\mathrm{min}(\bs{\theta}) = \mathbb{P} \big(\bs{\theta} \in \argmin_{\bs{\theta} \in \mathcal{D}} J(\bs{\theta}) \big).
	\label{eq:pmin}
\end{equation}
The approach becomes tractable by approximating~$p_\mathrm{min}$ on a non-uniform grid, with higher resolution in areas where it is more likely to find the minimum.
The key idea is that, upon convergence, we expect~$p_\mathrm{min}$ to be peaked around the minima, thus to have low entropy. The rate of change in the entropy of~$p_\mathrm{min}$ determines how much information about the location of the global minimum we obtain with each evaluation of the cost function. Given this metric, the optimal parameter at which to evaluate the cost function in the next iteration, is the one that is most informative:
\begin{equation}
	\bs{\theta}_{n+1} = \argmax_{\bs{\theta} \in \mathcal{D}} \E \left[  \Delta H(\bs{\theta}) \right],
	\label{eq:entropy_change}
\end{equation}
where~$\Delta H(\bs{\theta})$ is the change in entropy of~$p_\mathrm{min}$ caused by retrieving a new cost value at location~$\bs{\theta}$. 
Intuitively, by collecting cost values at the most informative locations~\cref{eq:entropy_change}, we keep decreasing the amount of entropy in $p_\mathrm{min}$ until eventually $p_\mathrm{min}$ is peaked around the optima.
At iteration $n$, we compute the best guess~$\bs{\theta}_\mathrm{bg}$ about the optimal parameters by minimizing the current GP posterior \cref{math:gp_prediction_mean}:
\begin{equation}
	\bs{\theta}_\mathrm{bg} = \argmin_{\bs{\theta} \in \mathcal{D}} \mu_n(\bs{\theta}).
\end{equation}

The computation of~$\Delta H$ given the GP model of~$J(\cdot)$ requires several approximations. A full derivation is beyond the scope of the paper, but all details can be found in~\cite{Hennig2012Entropy}.

\section{Reinforcement Learning with Simulations}
\label{sec:main_method}

In this section, we show how the ES algorithm can be extended to multiple sources of information, such as simulations and physical experiments. The two main challenges are modeling the errors of the simulator in a principled way and trading off evaluation effort and information gain. In the following, we focus on the case where only one simulation is available for ease of exposition. However, the approach can easily be extended to an arbitrary number of information sources.

\subsection{GP Model for Multiple Information Sources}
\label{sec:gp_multi_src}

To model the choice between simulation and phisical experiment, we use a specific kernel structure that is similar to the one used in~\cite{Poloczek2016MultiInformation}.
The key idea is to model the cost on the real system as being partly explained through the simulator plus some error term.
That is,~$J(\bs{\theta}) = J_\simu(\bs{\theta}) + J_\error(\bs{\theta})$, where the true cost consists of the estimated cost of the simulation,~$J_\simu(\bs{\theta})$, and a systematic error term,~$J_\error(\bs{\theta})$. To incorporate this in the GP framework of \cref{sec:background}, we extend the parameter vector by an additional binary variable~$\delta$, which indicates whether the cost is evaluated in simulation (${\delta=0}$) or on the physical system (${\delta=1}$). Based on the extended parameter~$\mb{a} = (\bs{\theta}, \delta)$, we can model the cost by adapting the GP kernel to
\begin{equation}
  k(\mb{a}, \mb{a}') = k_\simu(\bs{\theta}, \bs{\theta}') + k_\delta(\delta, \delta') \, k_\error(\bs{\theta}, \bs{\theta}').
  \label{eq:multi_information_kernel}
\end{equation}
The kernels~$k_\simu(\cdot, \cdot)$ and~$k_\error(\cdot, \cdot)$ model the cost function on the simulator and its difference to the cost on the physical system, respectively. The kernel~${k_\delta(\delta, \delta') = \delta \delta'}$ is equal to one if both parameters indicate a physical experiment and zero otherwise.

\begin{figure}[tb]
\centering
\begin{subfigure}{\columnwidth}
  \includegraphics[width=\textwidth]{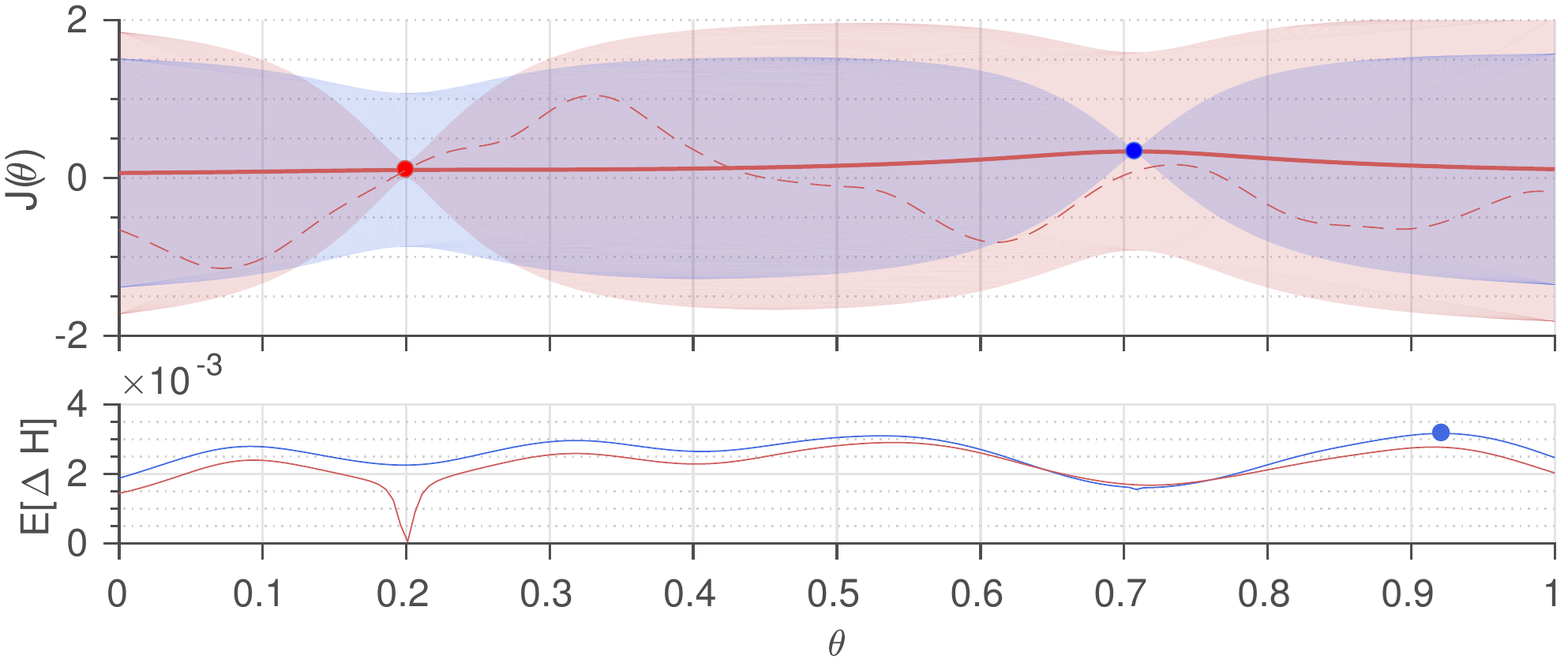}
  \caption{Exploration stage after two evaluations}
  \label{fig:synthetic_initial}
\end{subfigure} %
\begin{subfigure}{\columnwidth}
  \includegraphics[width=\textwidth]{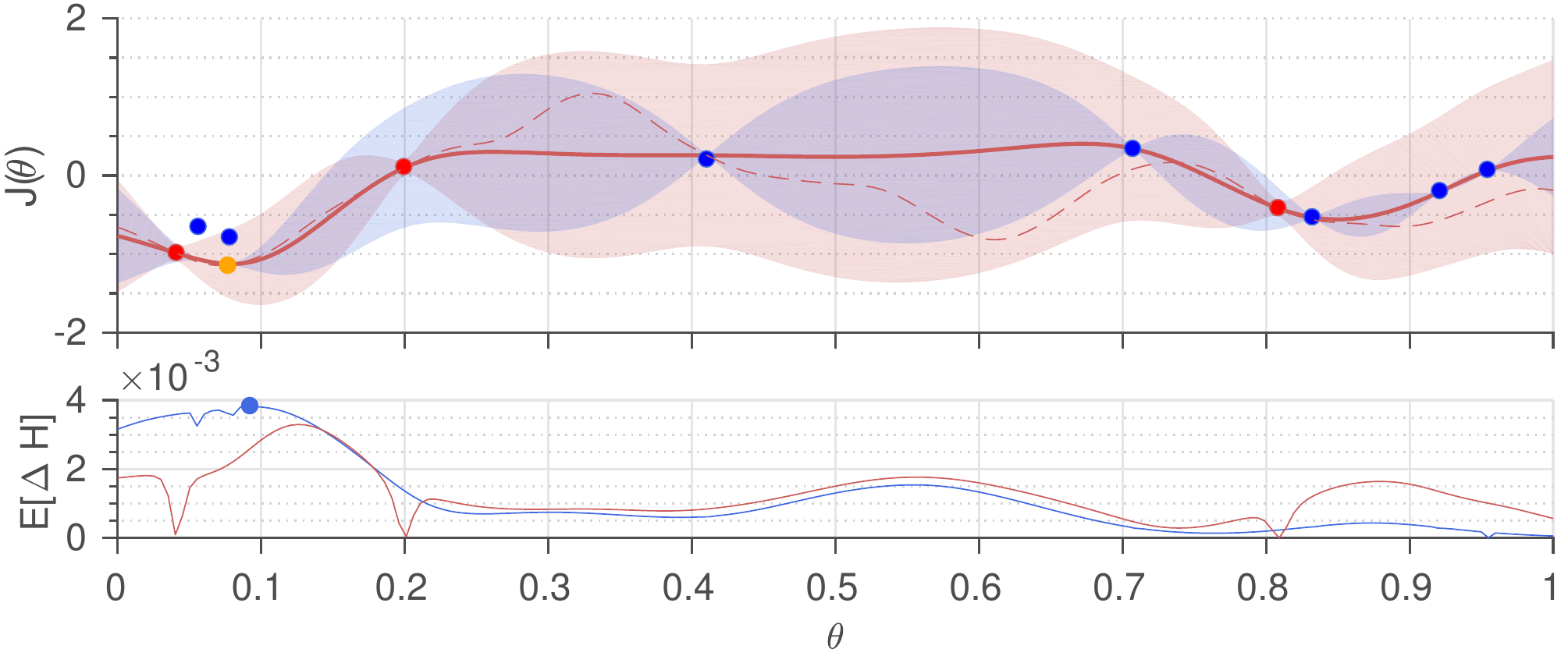}
  \caption{Exploration stage after ten evaluations}
  \label{fig:synthetic_final}
\end{subfigure} %
\caption{Synthetic example of how simulations and physical experiments can be combined by trading off information and evaluation effort. In (a), top, it is shown the GP posterior conditioned on one simulation (blue dot) and one physical experiment (red dot). The GP model from~\cref{sec:gp_multi_src} encodes that a portion of the uncertainty in the cost of the real system (red shaded) can be explained through the simulator (blue shaded). The red dashed line represents the cost function of the physical system. The cost function of the simulator is omitted for simplicity. In (a), bottom, it is shown the expected information gain per unit of effort of the simulator (blue line), and of the physical system (red line). The most informative point (blue dot) is selected among the two sources by the proposed method as next evaluation (in this case, a simulation). In (b), top, it is shown the GP posterior after nine iterations. The global minimum (orange dot) is found close to the true minimum.
}
\label{fig:synthetic_example}
\end{figure}

From \cref{sec:background}, we know that the kernel~\cref{eq:multi_information_kernel} models the covariances for different parameters. Intuitively, the kernel~\cref{eq:multi_information_kernel} encodes that two experiments on the physical system covary strongly. However, if one of the~$\delta$-variables is zero (i.e., a simulation), then the only covariances between the two values is captured by~$k_\simu$. Effectively, the error covariance is switched off in simulations in order to model that simulations cannot provide all the information about~$J$. By choosing the kernels $k_\simu$ and $k_\error$, we can model to what extend $J$ can be explained by the simulator and thereby its quality.
This is illustrated with a synthetic example in~\cref{fig:synthetic_example}. The total variance of the cost on the physical system is shown in red. 
Before any data is observed, it is equal to the uncertainty about the simulation plus the uncertainty about the error. As shown in~\cref{fig:synthetic_initial}, the blue shaded region highlights the variance of the simulator. Evaluations in simulation (blue dots) reduce the uncertainty of this blue shaded region, but reduce only partially the uncertainty about the true cost (red).
In contrast,
an evaluation on the real system (red dot) allows one to learn the true cost~$J$ directly, thus reducing the total uncertainty (red),
while some uncertainty about the variance of $J_\simu$ remains (blue). Having uncertainty about the simulation is by itself irrelevant for the proposed method, because we solely aim to minimize the performance on the physical system.

Next to the kernel, we account for different amounts of noise in simulation (typically noise-free) and on the real system. That is, the noise variance of measurements,~$\eta^2(\mb{a})$ in~\cref{eq:j_measurement}, takes different values, $\eta^2_\exper$ and $\eta^2_\simu$, depending on whether an experiment or a simulation is chosen.
With this kernel and noise structure, the two information sources can be modeled by a single GP, and predictions can be made according to~\cref{math:gp_prediction_mean,math:gp_prediction_variance}.

\subsection{Optimization}
\label{sec:optimization_multi_src}

With the GP model defined, we now consider how it can be used to trade off accuracy for evaluation effort.
As a first step, we quantify the goal. As before, we want to minimize the cost ~\cref{eq:cost_function} on the real system. This means, the distribution over the minimum is defined in terms of the same cost \cref{eq:pmin} as in standard ES. In order to approximate~$p_\mathrm{min}$, we need to use the GP kernel with the additional~$\delta$ factor fixed to one,
\begin{equation}
	p_\mathrm{min}(\bs{\theta}) = \mathbb{P} \big( \bs{\theta} \in \argmin_{\bs{\theta} \in \mathcal{D},\,\delta=1} J(\bs{\theta}, \delta) \big).
	\label{eq:pmin_multi}
\end{equation}

As in ES, the goal is to arrive at a distribution $p_\mathrm{min}$ that has low entropy (i.e., very peaked on a certain location). The expected change in entropy is an appropriate measure for this. However, this quantity additionally depends on the variable~$\delta$, so that the algorithm has an additional degree of freedom in the parameters to optimize. If one were to use the same optimization problem as in~\cref{eq:entropy_change}, the algorithm would always choose to evaluate parameters with~${\delta=1}$. This is because the experiments with~${\delta=1}$ provide information about the cost function~$J$ directly, while an evaluation with~${\delta=0}$ only provides information about part of the cost,~$J_\simu$.

To trade off between the two choices more appropriately, we associate an effort measure with both kinds of evaluations;~$t_\simu$ for the simulation and~$t_\exper$ for physical experiments. While simulations are less informative about~$p_\mathrm{min}$, they are significantly cheaper than experiments on a physical platform so that~${t_\simu < t_\exper}$. These effort measures can have physically meaningful units, such as the amount of time taken by a simulation relative to a physical experiment. While the effort measures are important to trade off the relative gains in information, they do not require tuning. For example, setting the effort of the simulator too high may lead to more experiments on the physical system than necessary, but the optimal parameters on the real system are found regardless.

A key advantage of using entropy to determine progress towards the goal is that it is a consistent unit of measurement for both information sources, even in the case of different noise variances. As a result, we can compare the gain in information about the location of the minimum (i.e.,~$p_\mathrm{min}$) in simulation and physical experiments relative to their efforts. Thus, we select the next parameters,~$\bs{\theta}_{n+1}$, and where to evaluate them, $\delta$, according to
\begin{equation}
  \argmax_{\bs{\theta} \in \mathcal{D},\, i \in \{\simu, \exper \}} \E\left[ \Delta H_i(\bs{\theta}) \right] \,/\, t_i.
  \label{eq:acquisition_function}
\end{equation}
The expected gain in entropy,~$\E\left[  \Delta H_i\right] $, depends on whether we evaluate in simulation or physical experiment. By selecting the best gain per unit of effort, the algorithm automatically decides which kind of evaluation decreases the uncertainty about the location of the minimum the most, relative to effort. Importantly, since the GP model in~\cref{eq:multi_information_kernel} is adaptive to the quality of the simulator, the acquisition function~\cref{eq:acquisition_function} leads to informed decisions about whether the simulator is reliable enough to lead to additional information.

We illustrate a typical run of the algorithm in~\cref{fig:synthetic_example}. The algorithm was initialized with one physical experiment (red dot in~\cref{fig:synthetic_initial}) for the purpose of illustration. The evaluation effort of the simulator was set to~40\% less of that of the real system.
As a result, it is advantageous to exploit initially the low effort that takes to do simulations. The algorithm automatically decides to do so, as can be seen in~\cref{fig:synthetic_initial}. The simulation (blue dot) decreases the amount of uncertainty about the simulation model, but provides only partial information about the true cost of the system. As a result,  the method eventually starts to evaluate parameters on the real system. Notice that this is not the same as two stage learning, because the algorithm can decide to switch back to simulations if this is beneficial. This is especially important in situations where the quality of the simulation is not known in advance and the hyperparameters of the kernels in~\cref{eq:multi_information_kernel} are optimized.
Eventually, the algorithm converges to a distribution~$p_\mathrm{min}$ that is peaked around the minima of the cost function.
Since the model can exploit cheap information from simulation, fewer physical experiments are needed to determine the minimum than if only physical experiments were used.

Because the proposed method extends Entropy Search  (ES) to multiple information sources, we refer to it as \emph{Multi-fidelity Entropy Search} (MF-ES). %

\section{Experimental Results}
\label{sec:experiments}

In this section, we evaluate \ours for optimizing the feedback controller of an unstable cart-pole system, as illustrated in  \cref{fig:concept}.

\subsection{Experimental Setup}
As experimental setup, we use the Quanser Linear Inverted Pendulum,~\cite{QuanserDoc}.
The dynamics of the system are described by
\begin{equation}
  \mb{x}_{k+1} = \mb{f}(\mb{x}_{k}, u_{k}),
  \label{eq:exp_dynamics}
\end{equation}
where $\mb{x}_{k} = [ s_k, \psi_k, \dot{s}_k, \dot{\psi}_k ]^\text{T} $ is the state at discrete time step~$k$, which is comprised of pendulum angle $\psi$, cart position $s$, and their time derivatives; $u_k$ is the commanded motor voltage driving the cart; and~$\mb{f}(\cdot)$ is the transition function (see \cite{QuanserDoc} for details).

The cart-pole setup is connected through dedicated hardware to a standard Laptop and can be controlled via Matlab/Simulink. A nonlinear Simulink model of the system dynamics \eqref{eq:exp_dynamics} is provided by the manufacturer and used as the simulator in our setting.

\subsection{Controller Tuning Problem}
\label{ssec:controller_tuning_problem}
To stabilize the pendulum about its upright equilibrium, we use a static state-feedback controller,
\begin{equation}
  u_k = \mb{F} \mb{x}_k,
  \label{eq:exp_stateFeedback}
\end{equation}
with gain matrix ${\mb{F} \in \mathbb{R}^{1 \times 4}}$.
We seek optimal gains $\bs{F}$ that minimize the cost function
\begin{equation}
J = \frac{1}{K} \sum_{k=0}^{K-1} s_k^2 + \psi_k^2 + \dot{s}_k^2 + 0.1 \dot\psi_k^2 + 10^{-1.5} u_k^2
\label{eq:exp_performanceCost}
\end{equation}
over a sufficiently long time horizon $K$.  The cost \eqref{eq:exp_performanceCost} penalizes deviations from the equilibrium $\mb{x} = 0$ and control effort ($u_k^2$).

Instead of tuning the controllers gains $\mb{F}$ directly (i.e. setting $\bs{\theta} = \mb{F}$), we follow the approach from~\cite{Marco2016Automatic,Trimpe2014Selftuning}, and pre-structure suitable controllers gains by means of a Linear Quadratic Regulator (LQR, \cite{Anderson2007Optimal}) design using a nominal, \emph{linearized} version, $(\mb{A},\mb{B})$, of the dynamics in~\eqref{eq:exp_dynamics}, around the aforementioned equilibrium, which can be obtained from the simulator, for example.  That is, the controller gain is computed from a discrete-time LQR design,
\begin{equation}
\mb{F} = \mathrm{dlqr}(\mb{A},\mb{B},\mb{W}_\mathrm{x}(\bs{\theta}), \mb{W}_\mathrm{u}(\bs{\theta}) ),
\label{eq:exp_lqrParameterization}
\end{equation}
where $\mb{W}_\mathrm{x}(\bs{\theta})$ and $\mb{W}_\mathrm{u}(\bs{\theta})$ are suitable parameterizations of LQR weights (see \cite{Marco2016Automatic,Trimpe2014Selftuning} for details).  Here, we selected
\begin{align}
\mb{W}_\mathrm{x}(\bs{\theta}) &= \diag (10^{\theta_1}, 1, 1, 0.1), && \theta_1 \in [-3, 2] \label{eq:exp_Wx},  \\
\mb{W}_\mathrm{u}(\bs{\theta}) &= 10^{-\theta_2}, && \theta_2 \in [1, 5]. \label{eq:exp_Wu}
\end{align}
Hence, we are left with tuning two parameters, ${\bs{\theta} \in \mathbb{R}^2}$.

Advantages of the LQR parameterization in~\eqref{eq:exp_lqrParameterization} are the possibility of dimensionality reduction, exploitation of prior knowledge in form of a linear dynamics model, and guarantees of stability and certain robustness properties with respect to the nominal system (see \cite{Marco2016Automatic} for a discussion).
However, we emphasize that LQR-weights are only one possible way to parameterize feedback controllers; alternative parameterizations~\cite{Roberts2011Feedback} or direct tuning of the gains $\mb{F}$~\cite{Berkenkamp2016Bayesian} is also possible. The method proposed herein is independent of the specific parameterization used.

With the above definitions, the cost function $J(\bs{\theta})$, which we seek to minimize \eqref{eq:cost_function}, is defined by equations \cref{eq:exp_performanceCost,eq:exp_lqrParameterization,eq:exp_Wx,eq:exp_Wu}. An evaluation of the cost~$J_\exper(\bs{\theta}^*)$ is obtained by computing the controller gain \eqref{eq:exp_lqrParameterization} based on the weight matrices \eqref{eq:exp_Wx}, \eqref{eq:exp_Wu}, performing a \unit[30]{s} balancing experiment on the physical system, and computing the cost according to \eqref{eq:exp_performanceCost} from the experimental data $\{\mb{x}_k, u_k \}_{k=0,\dots,K-1}$.
A simulation sample $J_\simu(\bs{\theta}^*)$ is obtained in the same manner by running a \unit[30]{s} simulation instead.

If a candidate controller violates safety limits on the states and inputs, it is determined as \emph{unstable}, and we assign a fixed penalty of ${J_\exper = 0.06}$ and ${J_\simu = 0.04}$ for physical experiment and simulation, respectively. These numbers are chosen conservatively larger than the cost of the worse stabilizing controller observed after some a priori initial evaluations. Thus, evaluations during the learning procedure shall not result in higher costs than these.

The controller is automatically tuned over roll-outs without human intervention. 
To this end, a nominal\footnote{The nominal controller is the optimal controller if the true dynamics was linear according to the nominal model $(\mb{A},\mb{B})$. Then, choosing $\mb{W}_\mathrm{x}(\bs{\theta})$ and $\mb{W}_\mathrm{u}(\bs{\theta})$ corresponding to the cost \eqref{eq:exp_performanceCost} yields the optimal controller $\mb{F}$, see~\cite{Marco2016Automatic}. This is a typical choice when neglecting the nonlinear dynamics.} controller $\bs{\theta}_\mathrm{nom} = \left[ 0,\: 1.5 \right]$ is balancing the pole when no tuning experiment is being performed. The optimizer triggers new experiments, when an evaluation on the real system is required. As soon as the experiment is finished, or instability is detected, the system switches back to the nominal controller.
The nominal controller shows very poor performance, which shall be improved with the proposed RL method. 

\subsection{Bayesian Optimization Settings}
\label{ssec:bos}
We apply the method of~\cref{sec:main_method}, MF-ES, to optimize the experimental cost \eqref{eq:exp_performanceCost} by querying simulations and experiments. %
The efforts in~\eqref{eq:acquisition_function} correspond to the approximate times we need to wait until a simulation is computed and a physical experiment is performed, $t_\simu = \unit[1]{s}$ and $t_\exper = \unit[30]{s}$, i.e., simulations require 30 times less effort than physical experiments.

\begin{figure}[t]
\centering
\includegraphics[width=\columnwidth]{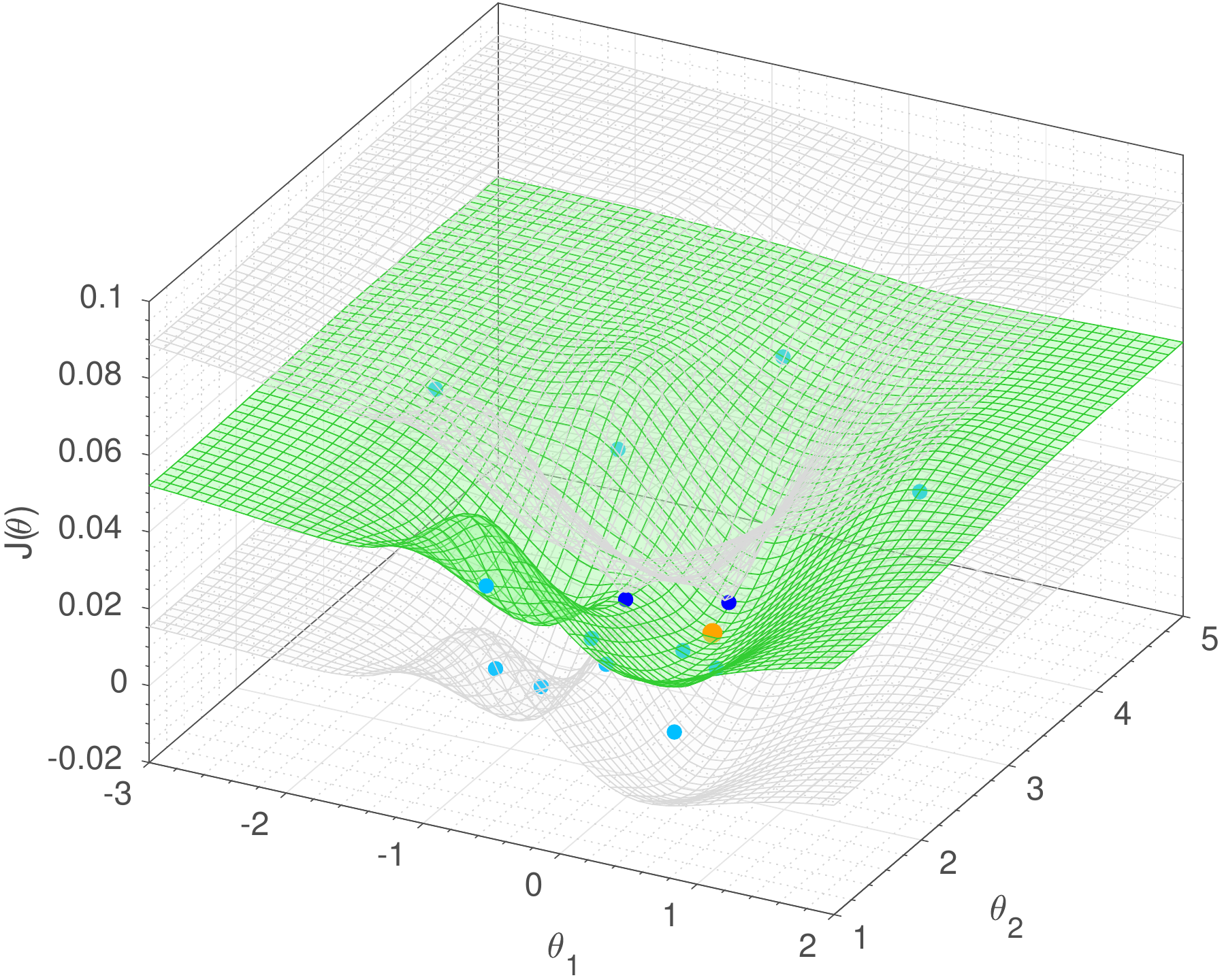}
\caption{GP posterior after termination of the exploration with MF-ES. The evaluations on the simulator (light blue) are systematically below the evaluations on the real system (dark blue). This bias is captured by the GP model assuming a lower prior mean for the simulator data, as mentioned in \cref{ssec:bos}. The posterior mean (green surface) and $\pm 2$ std (grey surface) predict the underlying cost function of the real system, conditioned on the observed data from both simulator and experiments. The best guess location for the global minimum, $\bs{\theta}_\text{bg}$, is represented by the orange dot.
}
\label{fig:GPexploration}
\end{figure}

For the GP model, we choose the rational quadratic kernel with~${\alpha = 1/4}$ (see \cite{Rasmussen2006Gaussian}) for both $k_\simu$ and $k_\error$ in \eqref{eq:multi_information_kernel}.  Hyperparameters, such as length scales and output variances, were chosen from some initial experiments and then held fixed during optimization.
As prior mean functions, we use ${m_\simu(\bs{\theta}) \equiv 0.04}$ and ${m_\error(\bs{\theta}) \equiv 0.02}$, respectively, for the simulation and error GP.  These choices
correspond to the penalties $J_\text{sim}$ and $J_\text{exp}$ given for unstable controllers (adding $m_\simu$ and $m_\error$ for the experiment).  Hence, the prior mean is pessimistic in the sense that we believe a controller to be unstable before seeing any data.
The prior variance of $k_\simu$ and $k_\error$ are chosen as $\sigma^2_\simu =1.6\times 10^{-5}$ and $\sigma^2_\error = 3.84\times 10^{-4}$ respectively.

The noise standard deviation of an evaluation on the real system, as defined in \eqref{eq:j_measurement}, has been estimated to ${\eta_\mathrm{exp} = 2.08\times 10^{-4}}$, while the noise of the simulator has been set to~${\eta_\mathrm{sim} = 10^{-5}}$, roughly twenty times lower.

We stop the exploration when the GP posterior mean at the best guess $\bs{\theta}_\mathrm{bg}$ (i.e., the current estimate of the global minimum) has not changed significantly (within a range of $\sigma_\error/4$ over the last 3 iterations), and we are sufficiently certain about its value (posterior standard deviation at $\bs{\theta}_\mathrm{bg}$ less than $\sigma_\error/2$). Once the exploration has terminated, we evaluate the final best guess controller on a physical experiment and take its cost as the outcome of the learning procedure.

\subsection{Results}

\begin{figure}[t]
\centering
\includegraphics[width=\columnwidth]{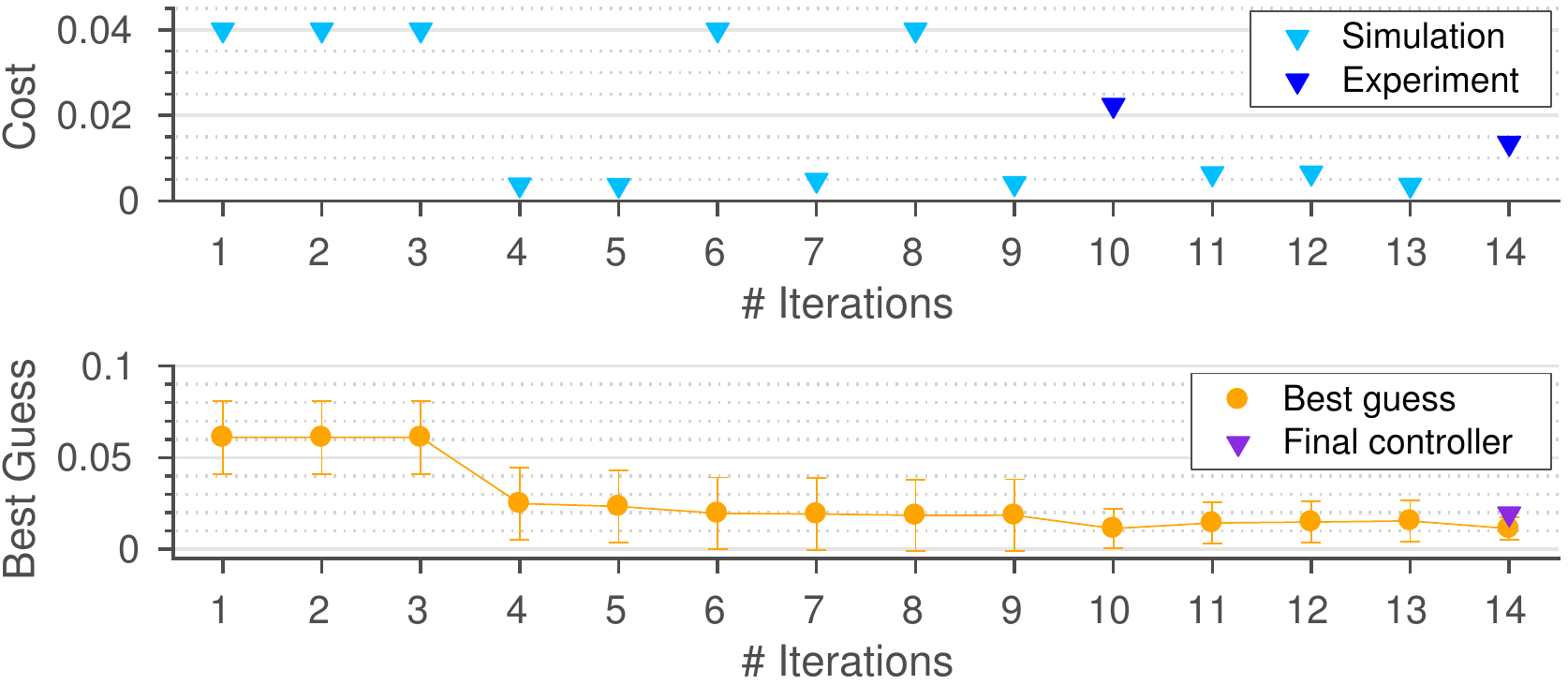}
\caption{(Top) Cost obtained at each iteration with the proposed approach during one exploration run. When the exploration terminates, the best guess is evaluated on the physical system (violet dot). (Bottom) Evolution of the GP posterior mean at the best guess $\mu_n(\bs{\theta}_\text{bg})$  and std $\pm \sigma_n(\bs{\theta}_\text{bg})$.}
\label{fig:performance}
\end{figure}

We run \ours on the LQR problem described in \cref{ssec:controller_tuning_problem}. \cref{fig:GPexploration} shows the final GP cost function landscape after the learning procedure, highlighting simulations (in light blue) and experiments (in dark blue). For the same learning run,~\cref{fig:performance} (top) illustrates how \ours alternates between simulations and physical experiments over iterations. As can be seen, the algorithm first performs multiple cheap simulations, which allow to identify regions of unstable controllers (i.e., regions of high predicted cost in \cref{fig:GPexploration}) without any real experiment. At iterations 10 and 14, the algorithm demands two expensive physical experiments. 
The reason is that a time unit spent in simulation is expected to be less informative than on a physical experiment. Thereby, experiment time should be better spent on the physical system. \cref{fig:performance} (bottom) shows the GP posterior mean and standard deviation of the best guess at each iteration. The stopping criterion terminates the exploration after 14 iterations because the GP posterior mean of the last three best guesses were steady enough. Finally, the algorithm selects the last global minimum, $\bs{\theta}_\mathrm{bg} = \left[  0.212,\: 2.42 \right] $ (orange dot in~\cref{fig:GPexploration}), as the final controller, which was evaluated on the physical system retrieving a low cost $J(\bs{\theta}_\mathrm{bg})=0.0194$.

As a remark, we observe that the algorithm alternates between simulations and experiments in a non-trivial way, which cannot be reproduced with a simple two-stage learning process, where simulations are used to seed experimental reinforcement learning. Furthermore, in \cref{fig:GPexploration}, we can see that the posterior mean around $\bs{\theta} = \left[ -2 , 4 \right] $ falls back to the prior in the absence of evaluations. As pointed out in \cref{ssec:bos}, the prior mean is pessimistic in the sense that predicts instability in unforeseen areas, which is a reasonable assumption in controller tuning of real systems.

\begin{figure}[tb]
\centering
\includegraphics[width=\columnwidth]{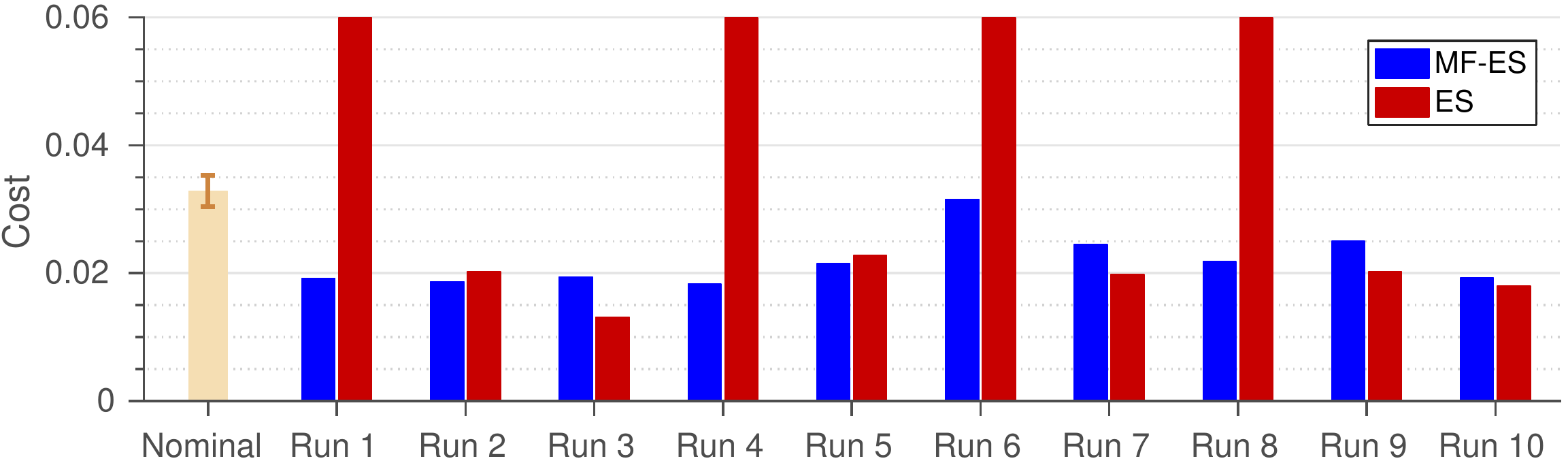}
\caption{Comparison of the final controller cost at each run between the proposed approach (MF-ES) and ES. The cost of the nominal controller (beige) with $\pm$ 2 std is shown for reference.}
\label{fig:cost}
\end{figure}

\begin{figure}[tb]
\centering
\includegraphics[width=\columnwidth]{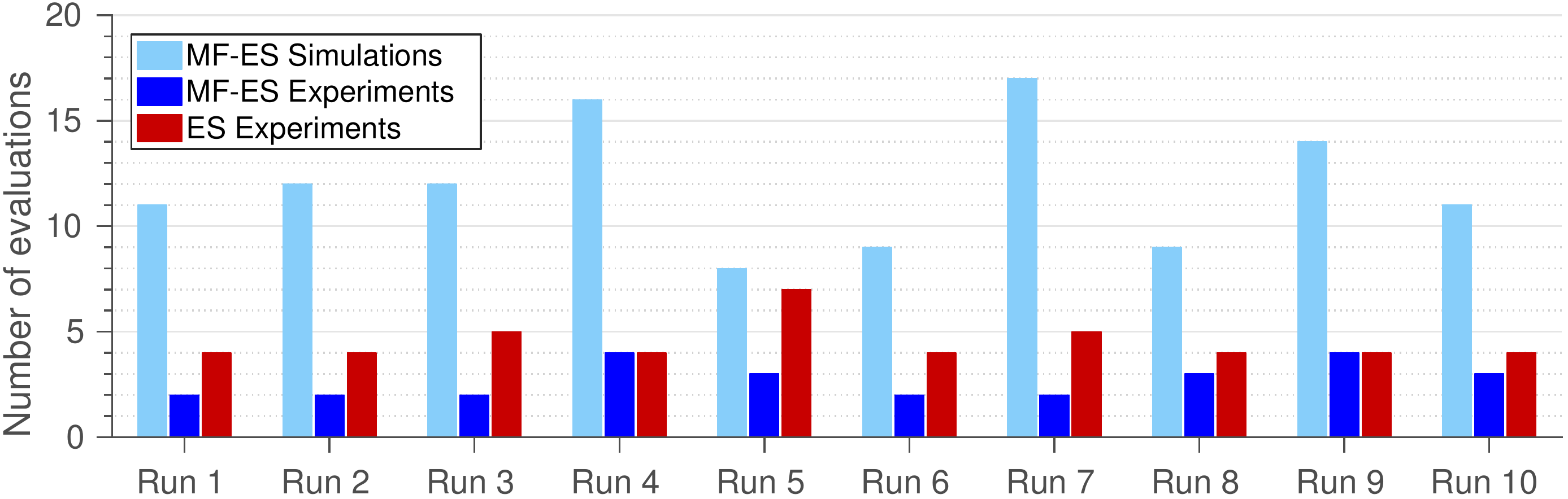}
\caption{Number of physical experiments at each run for \ours (dark blue) and standard ES (dark red), as well as simulations for \ours (light blue).}
\label{fig:n_exp}
\end{figure}

In order to
illustrate the benefit of trading off data from \emph{experiments and simulations},
we compare \ours to ES \cite{Hennig2012Entropy}, which uses \emph{only physical experiments}. The latter corresponds to the automatic controller tuning setting in \cite{Marco2016Automatic}. We run each of these methods ten times on the controller tuning problem. The results are discussed in \cref{fig:cost} and~\cref{fig:n_exp}.

In~\cref{fig:cost}, we show the cost of the final controller at each run, for both methods. The cost of the nominal controller (green) is shown as a reference. \ours finds controllers that are $33.23\%$ better, on average. Moreover, it consistently finds stabilizing controllers, while ES fails to find a stabilizing solution in 4 out of 10 cases (cost of 0.06).

\cref{fig:n_exp} compares the number of physical experiments performed with \ours (dark blue) and with ES (dark red). While ES needs on average $3.5$ physical experiments, \ours needs $2.7$ ($22.86\%$ less) plus $11.9$ simulations. These results demonstrate that \ours can find, on average, better controllers with a lower number of real experiments by also leveraging information from simulations.

\section{Conclusion}
\label{sec:conclusion}

We have shown a generic Bayesian optimization that can adaptively select between multiple information sources with different accuracies and evaluation efforts, such as experiments on a real robot and simulations. We applied this method to a policy optimization task on a cart-pole system. The experimental results confirm that using prior model information from a simulator can reduce the amount of data required to globally find good control policies.
 
\bibliographystyle{IEEEtran}
\bibliography{ICRA17_AMarco}

\end{document}